%% file: main.tex
\documentclass[pdflatex,sn-basic, iicol]{sn-jnl}% Math and Physical Sciences Numbered Reference Style
\usepackage{graphicx}%
\usepackage{booktabs}%
\usepackage{multirow}%
\usepackage{amsmath,amssymb,amsfonts}%
\usepackage{amsthm}%
\usepackage{mathrsfs}%
\usepackage[title]{appendix}%
\usepackage{xcolor}%
\usepackage{textcomp}%
\usepackage{manyfoot}%
\usepackage{algorithm}%
\usepackage{algorithmicx}%
\usepackage{algpseudocode}%
\usepackage{listings}%
\usepackage{url}%
\theoremstyle{thmstyleone}%
\theoremstyle{thmstyletwo}%

\theoremstyle{thmstylethree}%

\begin{document}

\title[Article Title]{iDocV2: Leveraging Self-Supervision and Open-Set Detection for Improving Pattern Spotting in Historical Documents}

%\title[Article Title]{Improving Speed Up by 10x on Pattern Spotting in Historical Documents through Self-Supervision and Open-Set Detection}


\author*[1]{\fnm{Jose M.} \sur{Saavedra}}\email{jmsaavedrar@miuandes.cl}

\author[1]{\fnm{Christopher} \sur{Stears}}\email{christopher.stears@ug.uchile.cl}

\author[1]{\fnm{Marcelo} \sur{Pizarro}}\email{marcelo.pizarro@ing.uchile.cl}

\author[1]{\fnm{Cristóbal} \sur{Loyola}}\email{cristobal.loyolam@usm.cl}

\author[1]{\fnm{Luis } \sur{Aros}}\email{laros1@uc.cl}

\affil*[1]{\orgdiv{Facultad de Ingeniería y Ciencias Aplicadas}, \orgname{Universidad de los Andes, Chile}, \orgaddress{\street{Av. Monseñor Álvaro del Portillo, 12455}, \city{Santiago}, \postcode{7620086}, \state{RM}, \country{Chile}}}


\abstract{Considering the imminent massification of digital books, it has become critical to facilitate searching collections through graphical patterns. Current strategies for document retrieval and pattern spotting in historical documents still need to be improved. State-of-the-art strategies achieve an overall precision of  $0.494$ for pattern spotting, where the precision for small non-square queries reaches 0.427. In addition, the processing time is excessive, requiring up to 7 seconds for searching in the DocExplore dataset due to a dense-based strategy used by SOTA models. Therefore, we propose a new model based on a better encoder (iDoc), trained under a self-supervised strategy, and an open-set detector to accelerate searching. Our model achieves competitive results with state-of-the-art pattern spotting and document retrieval, improving speed by 10x. Furthermore, our model reaches a new SOTA performance on the small non-square queries, achieving a new precision of 0.612.Different from the previous version, this leverages non-maximum suppression to reduce false positives. }




\maketitle

\input{01.introduction}
\input{02.related}
\input{03.proposal}

\input{04.settings}
\input{05.results}
\input{conclusions}

\section*{Declarations}
\begin{itemize}
\item Funding (information that explains whether and by whom the research was supported):  ANID, Chile.  Grant number ID23i10107

\item Conflicts of interest/Competing interests (include appropriate disclosures): Not applicable
\item Availability of data and material (data transparency): The data that support the findings is available at \url{https://www.docexplore.eu}.
\item Code availability (software application or custom code): You can find the code at \url{https://github.com/mpizarrot/iDoc}.
\item Ethics approval (include appropriate approvals or waivers): Not applicable.
\end{itemize}

\bibliography{references}% common bib file


\end{document}

%% file: 01.introduction.tex
\section{Introduction}
The original historical documents are extremely valuable and often irreplaceable. They are a crucial collective memory component, representing a nation's history, culture, and identity \cite{glayston:2024}. Moreover, the information extracted from historical documents allows people to reconstruct genealogies and perform demographic studies \cite{gellaty:2015}. Historical documents include ancient handwritten texts, old printed or photostat copies, records preserved on microfilm, and engineering or architectural drawings in city, county, state, and business archives. Thus, it is essential to maintain the integrity of these documents as they provide crucial insights into the people, places, and events of ancient times.

Digitization is a standard process to preserve the integrity of historical documents. In addition, digitization allows documents to be available online, facilitating access to researchers, students, interested institutions and anyone worldwide, while preventing damage by improper manipulation \cite{girdhar:2024, corbelli:2016}. Many components are involved in the  document digitization process depending on the preservation or interaction level you expect from your documents. Scanning, cleaning, segmentation and content extraction are examples of some stages involved in the digitization process \cite{sotirova:2012}. 

Once documents are available through digital means, other challenges arise. For instance, it is crucial to provide efficient mechanisms to query the content of digital documents. Traditional mechanisms use OCR or handwriting recognition algorithms to extract descriptions or annotations to facilitate keyword-based querying \cite{vilkomir:2024}. However, illustrations are a predominant type of content in medieval documents \cite{backhouse:1998}. Thus, it is also essential to develop efficient models for searching databases based on image content to complement traditional text-based searching.

Regarding the relevance of illustrations in historical document processing, our main interest is to provide effective and efficient mechanisms for graphical object localization (a.k.a. 
 pattern spotting) in digital documents, where a simple image is used as a query. Current strategies for pattern spotting in historical documents still show poor behavior, especially for spotting small patterns. Indeed, a recent work \cite{curi:2022} proposes a dense correlation-based method, achieving an overall mAP of 0.494 in the pattern spotting task on the DocExplore dataset \cite{en:2017}, where the performance for small non-square queries reaches 0.427\footnote{The Curi's performance was obtained by running the original code kindly provided by the first author.}. Although this method outperforms previous works, it undergoes an excessive searching time due to the dense strategy, requiring up to 7 seconds per query. In addition, current proposals are not robust to variations in scale of the target patterns.

%SOTA strategies achieves a precision around $0.58$ for document retrieval \cite{en:2017}. In the case of localization (a.k.a pattern spotting) the performance is much worse, achieving a precision around $0.27$ \cite{ubeda:2020}, which limits current methods to be used into real scenarios.

Therefore, this work proposes an efficient model for one-shot image detection in historical documents, changing the dense matching approach used in the Curi's proposal \cite{curi:2022} by sparse matching. Thus, our model reduces the querying time by 10x, making it appropriate for practical scenarios like searching in large libraries. 

Our proposal consists of two main components: an offline open-set region detector and a highly semantic encoder appropriately adapted to the context of historical documents using self-supervised learning. Our models show competitive precision regarding current methods in the pattern spotting task, increasing the effectiveness for the small non-square queries to 0.588, which allows us to reach a new overall mAP of 0.620. In addition, for document retrieval, our model achieves an overall mAP of 0.789, statistically similar to the best known model but 10x faster.

%is competitive with SOTA mode beats current approaches, achieving a mAP of $0.71$ for document retrieval and $0.54$ for pattern spotting,  which represents an increment in $22\%$ for the first task and $79\%$ for the second task. Our model reduce te 15 secondes proposed in .. to 120 ms}

%Therefore, this work proposes a new model that significantly improves the precision of document retrieval and pattern spotting in historical documents. Our proposal relies on two main components: an open-set candidate region detector and a high-semantic encoder appropriately adapted to the context of historical documents. Our results show that our proposal beats current approaches, achieving a mAP of $0.71$ for document retrieval and $0.54$ for pattern spotting,  which represents an increment in $22\%$ for the first task and $79\%$ for the second task.
 
To facilitate the reading, we organize this work as follows: Section \ref{sec:related} describes the related work. Section \ref{sec:proposal} presents our proposal. Section \ref{sec:settings} describes the experimental settings. Section \ref{sec:results} presents the achieved results. Finally, Section \ref{sec:conclusions} describes the final remarks.  

%% file: 02.related.tex
\section{Related Work}
\label{sec:related}
\subsection{Searching on Historical Documents}
Considering the upcoming widespread use of digital books, we have seen diverse approaches for searching by content in digital historical documents repositories. This interest arises even before the deep learning era. For instance, in 2010, Zhu and Keogh \cite{zhu:2010}  proposed the first method for mining historical manuscripts. They used a local color histogram and a divide-and-conquer approach to retrieve and locate queries from a few books. Later, \cite{yarlagadda:10} incorporated the Histogram of Oriented Gradients (HoG) and a color histogram in the HSV space for image representation. \cite{dovgalecs:2013} proposed a BoVW image representation using densely sampled SIFT and a Longest Weighted Profile algorithm to impose spatial ordering in the visual words representation. 

In 2017, \cite{en:2017} proposed the DocExplore dataset to standardize evaluating the proposed models for searching historical documents. In addition, they proposed a complete system for searching images and locating small graphic objects in images of medieval documents \cite{en:2016}. Their work is based on a region proposal module using  BING (Binarized Normed Gradients) \cite{cheng:2019} and a bag-of-words approach using VLAD and Fisher-Vector aggregation strategies.

With the flourishing of deep learning, new approaches emerged, particularly for encoding the region of interest. In this vein, the proposals of \cite{ubeda:2019, ubeda:2020} were pioneers in applying deep learning encoders for document retrieval and pattern spotting. To this end, they use an FPN architecture \cite{lin:2017} to represent images on different scales. The use of multiscale representations allows them to deal with the size diversity of the target patterns. The FPN-based Ubeda's proposal beat En's models in the pattern spotting task, increasing the performance from  0.157 to 0.272 (see Table \ref{tab:sota_map_ps}). However, the results in document retrieval did not follow the same way, achieving a performance slightly worse than En's proposal (see Table \ref{tab:sota_map_dr}).

More recently, \cite{curi:2022} proposed a correlation-based model in the feature domain. Using a VGG-16 \cite{simonyan:2015} as the image encoder, this model computes a feature map for the query and each catalog document. The model then correlates the query and document embeddings. This approach achieves the best document retrieval and pattern spotting performance but undergoes critical weakness. Curi's model is based on a dense correlation, which produces excessive time consumption, requiring over 7 seconds\footnote{All experiments were run with the same hardware, including a GPU RTX A6000 Ada.} to search in the DocExplore catalog, which only contains 1447 pages. In addition, the correlation approach assumes that the size of the target pattern is identical to the query's size, which is a strong constraint that can limit its applicability. 
 
\begin{table}[ht!]
\centering
\resizebox{\linewidth}{!}{%
\begin{tabular}{c c c c c}
\hline 
\multirow{2}{*}{size} &
\multirow{2}{*}{aspect ratio} &
\multicolumn{3}{c}{Pattern Spotting}\\ 
\cmidrule{3-5}
& &  En 2016 & Úbeda 2020 & Curi 2022*  \\ \hline \hline
big & square        &  0.546 & 0.681 & 0.877   \\ \hline
small & square      &  0.102 & 0.546 & 0.828   \\ \hline
big & non-square    & 0.405 & 0.509 & 0.749 \\ \hline
small & non-square  & 0.149 & 0.214 & 0.427   \\ \hline
 \multicolumn{2}{c}{mAP (\cite{en:2017})}  &
 0.157 &  0.272 & 0.494\\ \hline
 \multicolumn{5}{l}{\tiny{* performance obtained from the available code.}} 
 
 % \hline \multicolumn{8}{l}{ {\small  *we later rename it $mAP_{query-based}$}}
 \end{tabular}
}
\caption{Performance of SOTA models for pattern spotting. The metric is computed according to En's proposal \cite{en:2017}. The Curi`s method is the most effective one but it is highly expensive due to the use of a dense-based correlation. }

\label{tab:sota_map_ps}
\end{table}

\begin{table}[ht!]
\centering
\resizebox{\linewidth}{!}{%
\begin{tabular}{c c c c c }
\hline 
\multirow{2}{*}{size} &
\multirow{2}{*}{aspect ratio} &
\multicolumn{3}{c}{Document Retrieval}      \\ \cmidrule{3-5}
 & &  En 2016 & Úbeda 2020 & Curi 2022*       \\ \hline \hline
big & square &  0.881 & 0.749 & 0.932        \\ \hline
small & square &  0.801 & 0.742 & 0.923      \\ \hline
big & non-square &  0.701 & 0.660  &  0.813  \\ \hline
small & non-square  &  0.535 & 0.459 & 0.777 \\ \hline
 \multicolumn{2}{c}{mAP (\cite{en:2017})} &  0.580 &  0.577 & 0.799\\ \hline
 \multicolumn{5}{l}{\tiny{* performance obtained from the available code.}} 
 % \hline \multicolumn{8}{l}{ {\small  *we later rename it $mAP_{query-based}$}}
 \end{tabular}
}
\caption{Mean Average Precision of SOTA models for historical document image retrieval. Here, again, Curi's correlation-based model beats the others.}
\label{tab:sota_map_dr}
\end{table}

Tables \ref{tab:sota_map_ps} and \ref{tab:sota_map_dr} show the performance, so far known,  for pattern spotting and document retrieval, respectively, in the DocExplore dataset \cite{en:2017}. In this vein, the work of \cite{curi:2022} stands out as the best method reported so far regarding effectivity for document retrieval and pattern spotting. However, this proposal is highly inefficient because it is based on a full local correlation between the query and the patches in which a document is split down. %Indeed, our experiments show that Curi's method requires over 7 seconds for querying in a small catalog with 1500 pages. 

%Furthermore, Curi's method is not robust to scale variations of the target patterns. It assumes that the target objects in the documents are the same size as the query, which is not a big deal in the  DocExplore set because it does not show high variation in size between queries and target regions. 

% Table \ref{tab:sota_map} shows the performance, so far known,  for document retrieval and pattern spotting in the DocExplore dataset \cite{en:2017}. The SOTA model for document retrieval corresponds to the proposal of En et al. \cite{en:2016}, published in 2016, which achieves a mAP of $0.58$. However,  in the case of pattern spotting, the best-known model is the FPN-based model proposed by  Ubeda et al. \cite{ubeda:2020}  in 2020, which reaches $0.272$. 

Moreover, Tables \ref{tab:sota_map_ps} and \ref{tab:sota_map_dr}  point out the performance of the models over four groups of queries regarding their aspect ratio and size, according to the proposal of \cite{ubeda:2020}. Thus, the query patterns are organized into two sizes, big and small, and two aspect ratios, square and non-square. According to the SOTA performance, all the models work worse on the small non-square queries, which cover $83\%$ of the total set of queries of the DocExplore dataset. Furthermore, Ubeda's and Curi's approaches achieve a low performance on non-square queries, regardless of the size, for pattern spotting. Non-square queries can be extremely thin in the smallest size, and their information can be lost due to the reduction factor in the final feature map. 

Higher performance are achieved on document retrieval. Here, Curi's method reaches an overall mAP of 0.799, which is highly superior to the 0.494 achieved by the same model for pattern spotting. This result makes sense because pattern spotting is a finer task where the goal is to localize a query's occurrence in a document precisely. Moreover, a high precision on pattern spotting correlates with high precision on document retrieval. %In addition, document retrieval strongly depend on detection of target patterns. Therefore a better patter spotting model implies a better document retrieval. Thus

More recently, \cite{assaker:2025} presented a learning-based method for pattern spotting based on the DETR detection method \cite{carion:2020_detr}. The Assaker's proposal reported an overall performance of 0.381 for pattern spotting and 0.672 for image retrieval, being both significantly inferior to Curi's method. Unfortunately, since there is no information reported across the four query categories described above, we cannot include Assaker's results in the SOTA performance tables.

\subsection{Better Visual Encoders}
We are undergoing enormous advances on computer vision through better image representation models based on self-supervised learning to leverage a vast amount of available visual content without requiring explicit labels. In this vein, DINO \cite{caron:2021, oquab2023dinov2}, iBOT \cite{zhou:2021}, and CLIP \cite{radford2021learning} are popular SOTA models used to encode visual content for diverse downstream tasks \cite{kirillov2023segany, Rombach:2022}. However, to maximize the performance of encoders in a specific domain, it is crucial to adapt them strategically to avoid the catastrophe-forgetting phenomenon. Fortunately, current adaptation models like LoRA (Low Rank Adaptation) \cite{lora2022} have been shown to work appropriately, particularly for LLMs. Thus, we can leverage modern adaptation mechanisms to improve representations in the context of historical documents.  

Furthermore, open-set detectors can benefit the pattern-spotting task as a region proposal mechanism. In this vein, \cite{liu2023grounding} proposed Grounding DINO extending DINO object detector to work in an open-set regimen leveraging LLM representations. 
Grounding DINO can detect arbitrary objects with human inputs, such as category names or referring expressions. The power of an open-set detector is highly valuable for pattern spotting because it can deliver a set of interest regions to facilitate comparison concerning an input query. Recently, \cite{elhajj2023} combine Grounding DINO \cite{liu2023grounding}, and SAM \cite{kirillov2023segany} to effectively extract regions of interest from historical documents.

%seen as a particular case of the object detection problem. Instead of traditional and well-studied closed-set object detectors \cite{carion:2020_detr}, pattern spotting should be regarded as an open-set detector, where a query, whose class has not been known during training,  defines the target objects. Thus, pattern spotting is also understood as a particular case of one-shot detection, where the term \textit{one-shot} comes from the required query as input. 

However, pattern spotting and retrieval require more than a high-recall region extractor. As we will later show in this work, a key component of a visual search engine is a highly semantic encoder that appropriately represents the visual content of queries and target documents.   Therefore, the contribution of this work is a novel model for pattern spotting of historical documents, including a special encoder trained under a self-supervised regimen to allow the model to increase performance for searching historical records. We leverage all the knowledge spread during the last few years to significantly enhance the mechanisms for searching in digital repositories, offering a hopeful outlook for our work in this field. Even without high complexity, our model shows superior behavior for small non-square patterns, achieving a gain of $16.1\%$ over the best model. Moreover, our proposal reduces the querying time by a large margin, passing from 7 seconds to 720 ms, which can be further be improved  by indexing structures as described by \cite{toselli:2024}. 

%Our proposal effectivelly combines the advantages of Grounding DINO with self-supervised encoder adaptation mechanisms to provide an effective model for querying historical documents through visual queries. Our results show a significant increment in precision with respect to previous methods. In image retrieval, we achieve an overall performance of 0.71 of mAP for document retrieval and 0.54 for pattern spotting, which make our proposal the new state-of-the-art in searching historical medieval documents using graphical queries.  %, while in pattern spotting, we achieve 52%, demonstrating the robustness and reliability of our model.   

%% file: 03.proposal.tex
\section{The Proposal}
\label{sec:proposal}
\begin{figure}[ht!]
    \centering
    \includegraphics[width=\linewidth]{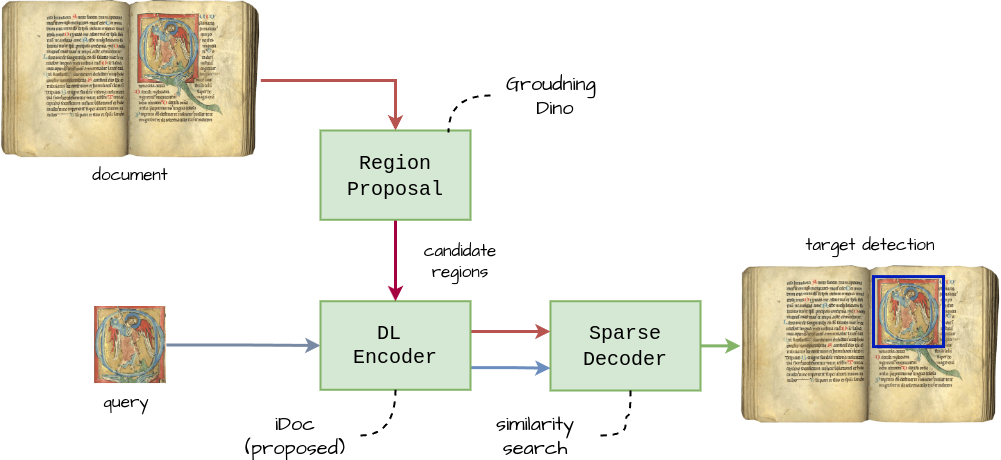}
    \caption{The general scheme of our proposal composed on a deep-learning encoder and similarity-search decoder. We also incorporate a semantic region proposal to increase speed up.}
    \label{fig:general-scheme}
\end{figure}

Figure \ref{fig:general-scheme} depicts the general scheme of our proposal. Like most modern machine learning models, ours consists of an encoder and a decoder module. The encoder extracts semantic information from the query and the target document, producing a set of feature vectors. To this end, we propose \textbf{iDoc}, a specific encoder for historical document patterns trained under a self-supervised strategy on the HORAE dataset. The decoder aims to match both sets of vectors, one obtained from the query and the other from each document, to spot occurrences of an input pattern. 
To speed up searching, we propose an offline region proposal module that transforms the documents into a set of regions of interest, which are then used for matching.. %Second, we include a method based on a topological-based dimensionality reduction to improve recall for small patterns. Following, we describe each of these components.

%Our general scheme of Figure \ref{fig:general-scheme}  may produce many instances of our general proposal by correctly instantiating the encoder and decoder with a specific architecture or model. In the following lines, we detail each component of our general proposal and the specific model for each. 

\subsection{Deep-Learning Encoder}
We instance the encoder with different SOTA visual representation models. However, we also propose a specific encoder for historical document graphical patterns leveraging a large dataset of medieval documents over which our encoder is trained using a self-supervised strategy.

\subsubsection{Pretrained Encoders}
We evaluate diverse visual encoders widely used in the computer vision community, Here, we selected the CLIP's visual encoder \cite{radford2021learning}, DINOv2 \cite{oquab2023dinov2, darcet2023vitneedreg} and iBOT \cite{zhou:2021} due to their demonstrated effectiveness on diverse downstream tasks.  %In addition, we evaluate a vanilla ResNet \cite{he:2016} trained on ImageNet to determine if a smaller and simpler model is competitive in our context concerning the performance of more elaborated attention-based models that exploit a huge amount of images by self-supervision.
     
\begin{itemize}
   \item \textbf{iBOT \cite{zhou:2021}}: this is a teacher-student network following the same architecture as BYOL \cite{grill:2020} or DINOv2 \cite{oquab2023dinov2} and trained with a non-contrastive loss. However, it uses patch-based masking modeling to allow the student to learn patch representations of masked regions as they were not masked. The backbone is a ViT similar to DINOv2 \cite{oquab2023dinov2}.
    
    \item \textbf{DINOv2}: this encoder consists of a ViT \cite{dosovitskiy:2021} that divides the image into a set of patches. It then computes an embedding for each patch together with a class token that absorbs global information. In our experiments, we used the ViT-B model with registers \cite{oquab2023dinov2, darcet2023vitneedreg}. This encoder was trained on a dataset of 142 million images by a self-supervised strategy. Moreover, the model uses an embedding size of 768 and a patch size of 14.
    \item \textbf{CLIP}: we use the ViT-B/16, pretrained on a dataset of 400 million image-text pairs \cite{radford2021learning}. The learned class token of 512-D is used as the final embedding. 
    %\item \textbf{ResNet}: we use a ResNet-50 pretrained on ImageNet-1k at resolution $224\times 224$. As the historical document context is widely different from the Imagenet dataset, we evaluate different residual blocks  instead of using just the last block.  To obtain the final representation, we add a global average pooling after the selected block.
\end{itemize}

\subsubsection{iDoc: the proposed encoder for historical documents trained by self-supervision}

We propose  \textbf{iDoc},  a new encoder specifically adapted for graphical patterns of historical documents. The proposed encoder relies on the following three aspects of design:

\begin{itemize}
    \item \textbf{A region extractor}:  splitting down each document into semantic pieces is critical to allow the encoder to be trained using coherent semantic regions. To this end, we use the Grounding DINO model discussed in Section \ref{sec:region_proposal}. As we will show later, using random regions produces a suboptimal performance compared to using semantic regions.
    
    \item \textbf{Self-supervision}: 
    self-supervision allows us to exploit a lot of visual data without requiring labeling. Considering the performance of different SOTA visual encoders presented in Table \ref{tab:baseline}, our training procedure is based on the iBOT model. 
    
    \item \textbf{Encoder adaptation}: current visual encoders have been trained with large datasets. Thus, keeping the learned knowledge while adapting it to a new context is crucial to produce a highly-semantic encoder. To this end,  we use the LoRA (\textbf{Lo}w-\textbf{R}ank \textbf{A}daptation)  adaptation strategy with the HORAE dataset using a pretrained iBOT model. 
\end{itemize}

\subsection{The Region Proposal}
\label{sec:region_proposal}
To reduce the comparison time, our model relies on a semantic region proposal applied offline to produce a set of candidate regions. Each region is then passed through the encoder to get a feature representation. We use Grounding DINO \cite{liu2023grounding} as the region extractor with the following set of prompts \texttt{figure - building - roof - sketch - person - symbol}.   The candidate regions are then filtered out by applying NMS with IoU=0.5, to avoid duplicates. The text prompts were based on the kind of objects of interest we can find in the DocExplore datasets. However, we try to use general concepts to avoid bias in specific objects. This set of terms can be changed according to the types of objects found in other historical document datasets.  

%We also evaluated the effect of varying the box and text thresholds on the obtained proposals. 

\subsection{The Sparse Decoder}

The encoder is based on a similarity search over the representations computed for the candidate regions and the input query. Here, we compute the cosine similarity between a query's encoding and the feature vectors obtained for all the regions extracted from the target documents. Finally, we sort all the proposals ascending  by their similarity score, obtaining the final ranking. %In addition, it is important to keep a link between a proposed region and the document it belongs to. Figure \ref{fig:sparse} depicts a scheme of our sparse-based decoder.

%% file: 04.settings.tex
\section{Experimental Settings} \label{sec:settings}
\subsection{The Dataset}
Our experiments involve two datasets: \textbf{DocExplore}, used mainly for evaluation and \textbf{HORAE} used only to train the proposed encoder. We provide more details of the involved datasets below:

\begin{itemize}
    \item \textbf{DocExplore} \cite{en:2017}: this dataset consists of manuscripts (referred to as pages) and symbols (referred to as queries) from the 10th to 16th centuries from the Municipal Library of Rouen, France. DocExplore consists of a total of 1500 pages and 1,447 graphical queries distributed in $35$ different categories. Figure \ref{fig:doc_explore_dataset} depicts examples of pages from the DocExplore dataset, including regions of interest extracted by Grounding DINO \cite{liu2023grounding}.%The pages were used for training and the queries for model evaluation. 

    \begin{figure}[ht!]
        \centering
        \includegraphics[width=\linewidth]{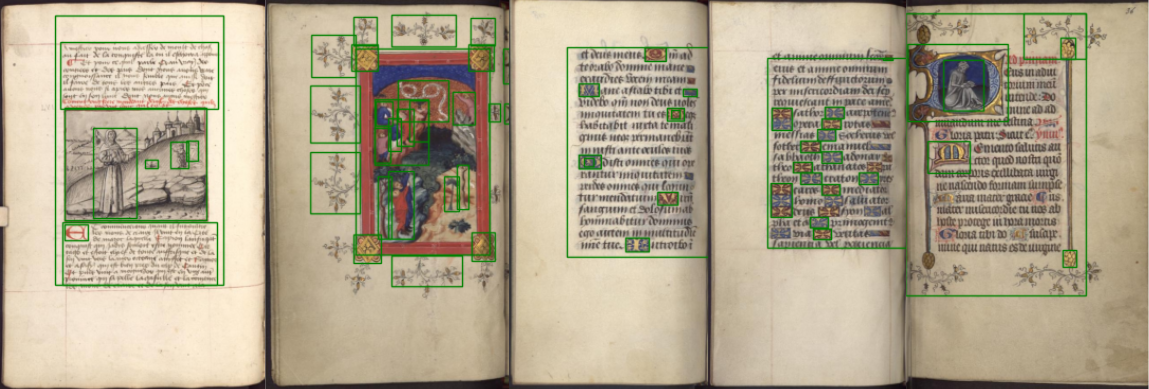}
        \caption{Pages from the DocExplore dataset with regions of interest extracted by Grounding DINO.}
        \label{fig:doc_explore_dataset}
    \end{figure}

    \item \textbf{HORAE} \cite{boillet2019horae}: this is a larger dataset that introduces a collection of annotated page from Books of Hours, a type of handwritten prayer book owned and used by rich lay people in the late Middle Ages \cite{boillet2019horae}. The corpus comprises 500 digitized manuscripts, containing a total of 107,227 images. Theses manuscripts were selected from a variety of digital libraries, primarily from France. Figure \ref{fig:horae-dataset}  depicts examples of pages from the HORAE dataset with region of interest extracted by Grounding DINO.

    For training the iDoc encoder,  $14$ manuscripts containing 12,947 pages were downloaded and processed with Grounding DINO using 0.01 as threshold. This process resulted in the extraction of 1,852,824 regions. From these, we randomly chose 270,000 regions for training.  
    \begin{figure}[ht!]
        \centering
        \includegraphics[width=\linewidth]{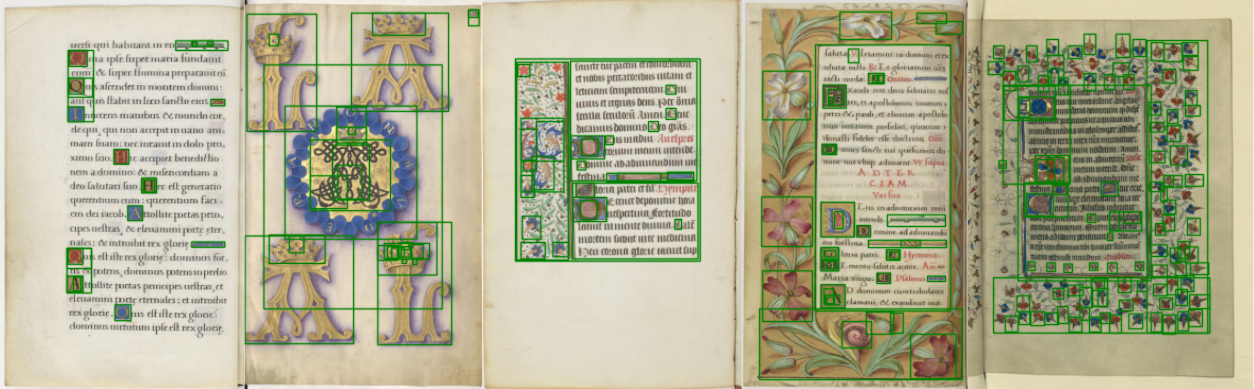}
        \caption{Pages from the HORAE dataset including regions of interest extracted by Grounding DINO.}
        \label{fig:horae-dataset}
    \end{figure}

\end{itemize}

\subsection{Defining the Experiments}
    %We run different experiments organized into the following two groups of experiments:
    As described in Figure \ref{fig:general-scheme}, our proposals are based on a sparse process based on extracting regions of interest by an open-set detection model.  An important benefit of incorporating a region proposal model is that it allows us to deal with occurrences in different scales, facilitating pattern spotting.  Using the Grounding DINO model for region extraction, we run the following experiments:
    
    \begin{itemize} 
    % \item \textbf{Evaluation of DINOv2 with a dense-based decoder}:
    % here, we are interested in evaluating modern encoders like DINOv2 \cite{oquab2023dinov2} in the context of document retrieval. These are simple experiments aiming to measure the capabitity of DINOv2 to encode patterns in historical documents. Thus, to this end, we only use a dense-based decoder.   
    
    \item \textbf{Evaluation of pretrained encoders}:
    we evaluate iBOT, DINOv2 and CLIP as visual encoders. We also include former visual encoders such as ResNet and VGG regarding the good results achieved by Curi's model using VGG.  
    
    \item \textbf{Evaluation of the LoRA-based adapted encoders}:
    we evaluate iDoc, our proposed encoder for historical documents. Here, we show the importance of using LoRA instead of adjusting all the parameters. In addition, we show the importance of using a semantic region proposal instead of a random selection. 
    
    %we adapt DINOv2-B by LoRA with different ranks ($r$). The same approach was applied to CLIP, but in this case using a maximum $r$ value of 512. Here, LoRA is applied to the attention blocks of each ViT, specifically to the Q, K and V matrices. In all our encoder adaptation experiments, we use the NT-Xent loss \cite{chen:2020}.  

    %We conduct experiments to show the performance of  our encoders after being adapted with HORAE and DocExplore datasets, independently. 

    % As we will see later, we evaluate our LoRA-based encoder adaptation using the HORAE dataset, a different set containing the target dataset, and using the DocExplore dataset. Using the HORAE dataset allow us to see the generalization capability of our models between different target catalogs.
    \end{itemize}
% \subsection{Settings}
% ??

\subsection{Parameter Settings}
We train iBOT on the HORAE training set with AdamW optimizer and a batch size of 64 for 100 epochs. The learning rate is linearly ramped up during the first 10 epochs to its base value scaled with the total batch size: lr $= 7.5e^{-4}\times batch\_size /256$. We use random masked image modeling (MIM), with prediction ratio r set as 0 with a probability of 0.5 and uniformly sampled from range [0.0, 0.7] with a probability of 0.5. For LoRA, we use a  rank of 64 for 50 epochs.

%% file: 05.results.tex
\section{Experimental Results}
\label{sec:results}

\subsection{Evaluation of SOTA visual encoders}
In this section, we describe our results of evaluating the incorporation of a sparse-based decoder using a region extractor component before the similarity search process. All the experiments use Grounding DINO as the region proposal model, where the prompt used to extract regions are:   \textit{figure},  \textit{building},  \textit{roof},  \textit{sketch},  \textit{person},  and \textit{symbol}. In addition, we set a confident threshold equal to $0.01$. 

According to our general scheme of Figure \ref{fig:general-scheme}, we instantiate the encoder with five common visual encoders in computer vision. Here, we evaluate iBOT \cite{zhou:2021}, DINOv2 \cite{oquab2023dinov2}, CLIP \cite{radford2021learning}, ResNet \cite{he:2016}, and VGG \cite{simonyan:2015}. The first three encoders use vision transformers, while the last two are convolutional-based encoders. We used the class token as the image representation for the encoder based on the vision transformer. Table \ref{tab:baseline} reports the results achieved by the pretrained encoders split according to the size and aspect ratios of the queries, following the same protocol of \cite{ubeda:2019}.

We observe the superiority of iBOT, which achieved an overall mAP of 0.433. In this vein, DINOv2 performs close to iBOT with 0.400 of mAP. This similar performance is related to the mAP achieved for the small non-square queries where both encoders produce similar results, 0.373 for iBOT and 0.363 for DINOv2.    

Furthermore, the non-square queries seem to be the more challenging patterns for all the evaluated encoders. In addition, the small non-square patterns are those where all the models achieve their worst performance. Here, iBOT and DINOv2 achieve 0.373 and 0.363, respectively, while VGG reaches just 0.134. This fact is also shown in Table \ref{tab:sota_map_ps}, where all the current proposals achieve the worst result for the small non-square queries.

\begin{table}[ht!]
\centering
\resizebox{\linewidth}{!}{%
\begin{tabular}{ c c  c c c c c }
\hline
\multicolumn{1}{c}{size} & \multicolumn{1}{c||}{aspect ratio} &
\multicolumn{1}{c}{iBOT} & 
\multicolumn{1}{c}{DINOv2} & 
\multicolumn{1}{c}{CLIP} & 
\multicolumn{1}{c}{VGG19} & 
\multicolumn{1}{c}{Resnet50}\\ \hline \hline
\multicolumn{1}{c}{big} & \multicolumn{1}{c||}{square} & 
\multicolumn{1}{c}{0.716} & 
\multicolumn{1}{c}{0.640} & 
\multicolumn{1}{c}{\textbf{0.722}} & 
\multicolumn{1}{c}{0.558} & 
\multicolumn{1}{c}{0.633}  \\ \hline
\multicolumn{1}{c}{small} & \multicolumn{1}{c||}{square} &
\multicolumn{1}{c}{\textbf{0.746}} & 
\multicolumn{1}{c}{0.581} & 
\multicolumn{1}{c}{0.567} & 
\multicolumn{1}{c}{0.640} & 
\multicolumn{1}{c}{0.652} \\ \hline
\multicolumn{1}{c}{big} & \multicolumn{1}{c||}{non-square} &
\multicolumn{1}{c}{\textbf{0.653}} &
\multicolumn{1}{c}{0.562} & 
\multicolumn{1}{c}{0.545} &
\multicolumn{1}{c}{0.430} & 
\multicolumn{1}{c}{0.533} \\ \hline
\multicolumn{1}{c}{small} & \multicolumn{1}{c||}{non-square} & 
\multicolumn{1}{c}{\textbf{0.373}} &
\multicolumn{1}{c}{0.363} &
\multicolumn{1}{c}{0.210} &
\multicolumn{1}{c}{0.134} & 
\multicolumn{1}{c}{0.231} \\ \hline \hline
\multicolumn{2}{c||}{$mAP_{query-based}$} & 
\multicolumn{1}{c}{\textbf{0.433}} &
\multicolumn{1}{c}{0.400} &
\multicolumn{1}{c}{0.273} &
\multicolumn{1}{c}{0.213} &
\multicolumn{1}{c}{0.299} \\ \hline

\end{tabular}
}
\caption{Performance on pattern spotting achieved by five common visual encoders. Here, we use Grounding DINO as region extractor. %detections obtained with prompt \texttt{figure. building. roof. sketch. person. symbol
}
\label{tab:baseline}
\end{table}

\subsection{iDoc: the new encoder for historical document patterns}
%In the previous sections, we presented results using SOTA visual encoders joined to a sparse decoder. However, the results do not beat the model proposed by En \cite{en:2016} for document retrieval nor the proposal of Ubeda \cite{ubeda:2020}  for pattern spotting. Hence, we can conclude that current visual encoder and a good region proposal model like Grounding DINO are insufficient to beat current models for searching in historical documents, even though an dense-based decoder is an better approach than a sparse one. 
Here, we go further, adapting the visual encoder to the medieval document context. To this end, we leverage the LoRA \cite{lora2022} mechanism to allow the adaptation of the information produced for the diverse attentional layers in the ViT used iBOT. 

We present the results after adapting iBOT using the LoRA strategy with the HORAE dataset. This allows us to measure the generalization power of our adapted encoder into a different dataset like DocExplore.  Thus, we propose a superior encoder reaching 0.620 of mAP in the DocExplore dataset, with an outstanding performance for small non-square queries. Our new visual encoder named \textbf{iDoc} achieves new state-of-the-art results of 0.588 of mAP for the small non-square queries, considered the most challenging patterns. 

Furthermore, in the next sections, we show the contribution of critical aspects of design, like using an adaptation mechanism for training instead of adjusting all the parameters and using a semantic region extractor instead of random crops. 
%In the next lines, we first discuss the performance of our HORAE-centered adapted models for the document retrieval task. Then, we discuss the performance of our proposal for pattern spotting. For all the experiments in this section, we present the performance results changing the LoRA rank ($r$) in the range  $\{ 32, 64, 128, 256, 512, 768\}$.
%Here, we present the results after adapting CLIP and DINOv2 using the LoRA strategy with the HORAE dataset. This allows us to measure the generalization power of our adapted encoder between two different datasets. In the next lines, we first discuss the performance of our HORAE-centered adapted models for the document retrieval task. Then, we discuss the performance of our proposal for pattern spotting. For all the experiments in this section, we present the performance results changing the LoRA rank ($r$) in the range  $\{ 32, 64, 128, 256, 512, 768\}$.

\subsubsection{The importance of LoRA-based  adaptation}

Table \ref{tab:ibot_lora} shows the results comparing our proposal iDoc (iBOT + LoRA) and the version trained from random initialization. We observe that training from scratch is very competitive with respect to using LoRA. However, the best results are achieved with the adaptation mechanism, particularly for the big non-square queries where our proposal increases the precision by $7\%$,   producing an overall improvement of $2\%$ in the overall mAP. It suggests that the domain where iBOT was pretrained is very different from our domain, leveraging a small part of the learned knowledge.    

Of course, we can also try with a finetuning, adjusting all the parameters from the iBOT weights, but we do not include it for the following reasons:

\begin{enumerate}
    \item There is vast evidence about the catastrophic forgetting effect when adjusting all the pretrained parameters in a new domain \cite{chen:2025_forget, jiang:2025_forget}. 
    
    \item iBOT consists of 85,798,656 parameters and LoRA needs only 3,529,728, representing 4.11\% of iBOT. Thus, adapting by LoRA allows us to reduce the training time largely, producing models quickly. This fact is crucial in environments with access to limited computing power.
\end{enumerate}

%% Pattern spotting
\begin{table}[ht!]
\centering
\resizebox{\linewidth}{!}{%
\begin{tabular}{c c c c}
\hline
% \multicolumn{5}{|c|}{Adapted on HORAE dataset (r16) --- o r64}\\ \hline
\multicolumn{1}{c}{size} & \multicolumn{1}{c||}{aspect ratio} & 
\multicolumn{1}{c}{iDoc (iBOT + LoRA)} &
\multicolumn{1}{c}{iBOT from Scratch}  \\ \hline \hline
\multicolumn{1}{c}{big} & 
\multicolumn{1}{c||}{square} & 
\multicolumn{1}{c}{\textbf{0.786}} &
\multicolumn{1}{c}{0.773} \\ \hline
\multicolumn{1}{c}{small} & 
\multicolumn{1}{c||}{square} &
\multicolumn{1}{c}{\textbf{0.798}} &
\multicolumn{1}{c}{0.780}  \\ \hline
\multicolumn{1}{c}{big} & 
\multicolumn{1}{c||}{non-square} &
\multicolumn{1}{c}{\textbf{0.641}} &
\multicolumn{1}{c}{0.565} \\ \hline
\multicolumn{1}{c}{small} & 
\multicolumn{1}{c||}{non-square} & 
\multicolumn{1}{c}{\textbf{0.588}} &
\multicolumn{1}{c}{0.570}   \\ \hline \hline
\multicolumn{2}{c||}{$mAP_{query-based}$} & 
\multicolumn{1}{c}{\textbf{0.620}} &\multicolumn{1}{c}{0.602}   \\ \hline

\end{tabular}}

\caption{A comparison of our proposed encoder when iBOT is trained using LoRA (iDoc) over the pretrained version versus when it is trained from scratch. The results show the effectiveness of using adaptation by LoRA, allowing the models to improve in $7\%$ the mAP for the big non-square queries.}

\label{tab:ibot_lora} 
\end{table}

\subsubsection{The importance of semantic regions}
A key aspect of our proposal is using a semantic region extractor to build the set of regions of interest for training. Table \ref{tab:ibot_lora_random} presents the results comparing our proposal using Grounding DINO  as a region extractor against using random regions. Note that the use of random regions is a generalization of using a homogeneous split because, in a regular split,  each cell can contain a random section. In addition, random regions can reduce bias by allowing a larger dataset of regions.   Our experimental results show the superiority of our proposal using semantic regions, improving the mAP from 0.448 with random regions to 0.620 using Grounding DINO, representing a gain of $17.2\%$.

%almost doubling the precision with respect the random version. Here, our models acheives  that achieve a mAP $32.3\%$.

\begin{table}[ht!]
\centering

\resizebox{\linewidth}{!}{%
\begin{tabular}{c c c c}
\hline
% \multicolumn{5}{|c|}{Adapted on HORAE dataset (r16) --- o r64}\\ \hline
\multicolumn{1}{c}{size} & 
\multicolumn{1}{c||}{aspect ratio} & 
\multicolumn{1}{c}{iDoc (iBOT+LoRA+GD)} &
\multicolumn{1}{c}{iBOT+LoRA+random}  \\ \hline \hline
\multicolumn{1}{c}{big} &
\multicolumn{1}{c||}{square} & 
\multicolumn{1}{c}{\textbf{0.786}} &
\multicolumn{1}{c}{0.684} \\ \hline
\multicolumn{1}{c}{small} &
\multicolumn{1}{c||}{square} &
\multicolumn{1}{c}{\textbf{0.798}} &
\multicolumn{1}{c}{0.710}  \\ \hline
\multicolumn{1}{c}{big} &
\multicolumn{1}{c||}{non-square} &
\multicolumn{1}{c}{\textbf{0.641}} &
\multicolumn{1}{c}{0.540} \\ \hline
\multicolumn{1}{c}{small} &
\multicolumn{1}{c||}{non-square} & 
\multicolumn{1}{c}{\textbf{0.588}} &
\multicolumn{1}{c}{0.400}   \\ \hline \hline
\multicolumn{2}{c||}{$mAP_{query-based}$} & 
\multicolumn{1}{c}{\textbf{0.620}} &
\multicolumn{1}{c}{0.448}   \\ \hline

\end{tabular}}

\caption{A comparative study between our proposal iDOC based on Grounding DINO versus a version trained with random regions. Our model increases the overall mAP in $17.2\%$.}
\label{tab:ibot_lora_random}
\end{table}

\subsubsection{Class token versus aggregation of  patch encoding}
We also evaluate the importance of using the class token to encode the input image instead of aggregating the patch representations. Here, we use the average over the patch representations of the input image to compute a patch-based encoding. Table \ref{tab:ibot_lora_patch} presents the results of this evaluation, showing that the use of class token reaches a gain of $10\%$ over the patch-based aggregation.  

\begin{table}[ht!]
\centering
\resizebox{\linewidth}{!}{%
\begin{tabular}{ c c c c }
\hline
% \multicolumn{5}{|c|}{Adapted on HORAE dataset (r16) --- o r64}\\ \hline
\multicolumn{1}{c}{size} &
\multicolumn{1}{c||}{aspect ratio} & 
\multicolumn{1}{c}{iDoc (iBOT+LoRA+GD)} &
\multicolumn{1}{c}{iBOT+patch-avg}  \\ \hline \hline
\multicolumn{1}{c}{big} &
\multicolumn{1}{c||}{square} & 
\multicolumn{1}{c}{\textbf{0.786}} &
\multicolumn{1}{c}{0.6507} \\ \hline
\multicolumn{1}{c}{small} &
\multicolumn{1}{c||}{square} &
\multicolumn{1}{c}{\textbf{0.798}} &
\multicolumn{1}{c}{0.7455}  \\ \hline
\multicolumn{1}{c}{big} & \multicolumn{1}{c||}{non-square} &
\multicolumn{1}{c}{\textbf{0.641}} & 
\multicolumn{1}{c}{0.5504} \\ \hline
\multicolumn{1}{c}{small} &
\multicolumn{1}{c||}{non-square} & 
\multicolumn{1}{c}{\textbf{0.588}} & 
\multicolumn{1}{c}{0.4820}   \\ \hline \hline
\multicolumn{2}{c||}{$mAP_{query-based}$} & 
\multicolumn{1}{c}{\textbf{0.620}} &
\multicolumn{1}{c}{0.5206}   \\ \hline

\end{tabular}}

\caption{A comparative study between using  class token or a patch-based aggregation. Our results show the benefit of using the first one with a precision increment of $10\%$ with respect to a patch-based aggregation.}
\label{tab:ibot_lora_patch}
\end{table}

\subsubsection{Time Processing}
Our proposed model shows a speed-up of 10x over Curi's, improving the precision of small non-square queries by $16.1\%$ (gain of $37.7\%$). Our model requires 720 ms to search patterns in the DocExplore dataset, while Curi's proposal needs around 7 seconds — all the experiments were run on the same hardware using a GPU RTX A6000 Ada.

In terms of the number of operations, Curi's method needs to correlate the query feature map over each page feature map. Considering that the average page size is $602 \times 920$ and the visual encoder is affected by a stride of 5x, the size of the page feature map results in $120 \times 184$ vectors. Consequently, the Curi's approach requires 22,080 comparisons with the query feature map. In contrast, our method detects, on average, 112 regions per page, which needs 100x less comparison than Curi's. 
%takes 18 min to process all the queries which ... Therefore the processing time for each query is 70 ms, 100x faster than the Curi's method that overpass 7 minutes per query. % iDoc takes 23 min to process all the queries. Therefore, the oricessubg tune for each query is 953ms

%\subsubsection{Performance by query class}
\subsubsection{Document Retrieval}
Table \ref{tab:ps} summarizes the performance of our model using iDoc against Curi's model, where we highlight the outstanding performance of our proposal for small, non-square patterns. However, beyond pattern spotting, we present comparative results on the document retrieval task following the evaluation protocol of \cite{en:2016}. Table \ref{tab:dr} illustrates the competitive performance of our model for all the query categories, achieving an overall map of 0.789, just 0.01 of a difference from Curi's result. Furthermore, after running a statistical test, in this case the U Mann-Whitney test, given the non-normality of the results, the obtained $p$-value is $0.47$, showing that both methods perform very similarly to each other.

\begin{table}[ht!]
\centering

\resizebox{\linewidth}{!}{%
\begin{tabular}{ c c c c c c c }
\hline
\multicolumn{7}{c}{Pattern Spotting} \\ \hline 
\multicolumn{1}{c}{size} & 
\multicolumn{1}{c||}{aspect ratio} & 
\multicolumn{1}{c}{iDoc (v1)} &
\multicolumn{1}{c}{Curi's model} &
\multicolumn{1}{c}{iDoc-nms} &
\multicolumn{1}{c}{iDoc-mix} &
\multicolumn{1}{c}{iDoc-mix-nms} \\ \hline \hline

\multicolumn{1}{c}{big} & 
\multicolumn{1}{c||}{square} & 
\multicolumn{1}{c}{0.786} &
\multicolumn{1}{c}{0.877} &
\multicolumn{1}{c}{0.868} &
\multicolumn{1}{c}{0.784} &
\multicolumn{1}{c}{0.860} \\ \hline

\multicolumn{1}{c}{small} &
\multicolumn{1}{c||}{square} &
\multicolumn{1}{c}{0.798} &
\multicolumn{1}{c}{0.828} &
\multicolumn{1}{c}{0.813} &
\multicolumn{1}{c}{0.807} &
\multicolumn{1}{c}{0.821} \\ \hline

\multicolumn{1}{c}{big} &
\multicolumn{1}{c||}{non-square} &
\multicolumn{1}{c}{0.641} &
\multicolumn{1}{c}{0.749} &
\multicolumn{1}{c}{0.679} &
\multicolumn{1}{c}{0.624} &
\multicolumn{1}{c}{0.652} \\ \hline

\multicolumn{1}{c}{small} & 
\multicolumn{1}{c||}{non-square} & 
\multicolumn{1}{c}{0.588} &
\multicolumn{1}{c}{0.427} &
\multicolumn{1}{c}{0.602} &
\multicolumn{1}{c}{0.599} &
\multicolumn{1}{c}{0.612} \\ \hline \hline

\multicolumn{2}{c||}{$mAP_{query-based}$} & 
\multicolumn{1}{c}{0.620} & 
\multicolumn{1}{c}{0.494} &
\multicolumn{1}{c}{0.637} &
\multicolumn{1}{c}{0.630} &
\multicolumn{1}{c}{0.645} \\ \hline

\end{tabular}
}
\caption{Performance of iDoc on the pattern spotting task compared with Curis' model. Although Curi's model performs better for big and square queries, our proposal is particularly effective for small, non-square patterns.}

\label{tab:ps}
\end{table}

\begin{table}[ht!]
\centering
\resizebox{\linewidth}{!}{%
\begin{tabular}{ c c c c c }
\hline
\multicolumn{5}{c}{Document Retrieval} \\ \hline 
\multicolumn{1}{c}{size} & 
\multicolumn{1}{c||}{aspect ratio} & 
\multicolumn{1}{c}{iDoc (v1)} &
\multicolumn{1}{c}{Curi's model} &
\multicolumn{1}{c}{iDoc-mix} \\ \hline \hline

\multicolumn{1}{c}{big} & 
\multicolumn{1}{c||}{square} & 
\multicolumn{1}{c}{0.923} &
\multicolumn{1}{c}{0.932} &
\multicolumn{1}{c}{0.917} \\ \hline

\multicolumn{1}{c}{small} &
\multicolumn{1}{c||}{square} &
\multicolumn{1}{c}{0.906} &
\multicolumn{1}{c}{0.923} &
\multicolumn{1}{c}{0.908}  \\ \hline

\multicolumn{1}{c}{big} &
\multicolumn{1}{c||}{non-square} &
\multicolumn{1}{c}{0.800} &
\multicolumn{1}{c}{0.813} &
\multicolumn{1}{c}{0.789} \\ \hline

\multicolumn{1}{c}{small} & 
\multicolumn{1}{c||}{non-square} & 
\multicolumn{1}{c}{0.767} &
\multicolumn{1}{c}{0.777} &
\multicolumn{1}{c}{0.776} \\ \hline \hline

\multicolumn{2}{c||}{$mAP_{query-based}$} & 
\multicolumn{1}{c}{0.789} & 
\multicolumn{1}{c}{0.799} &
\multicolumn{1}{c}{0.796} \\ \hline

\end{tabular}
}
\caption{Performance of our model iDoc on the document retrieval task. Our results are very close to the results achieved by Curi's model ($p$-value = 0.47). However, ours is 10x faster.}

\label{tab:dr}
\end{table}

\subsubsection{Qualitative results}
\begin{figure}[ht!]
     \centering
     \includegraphics[width=\linewidth]{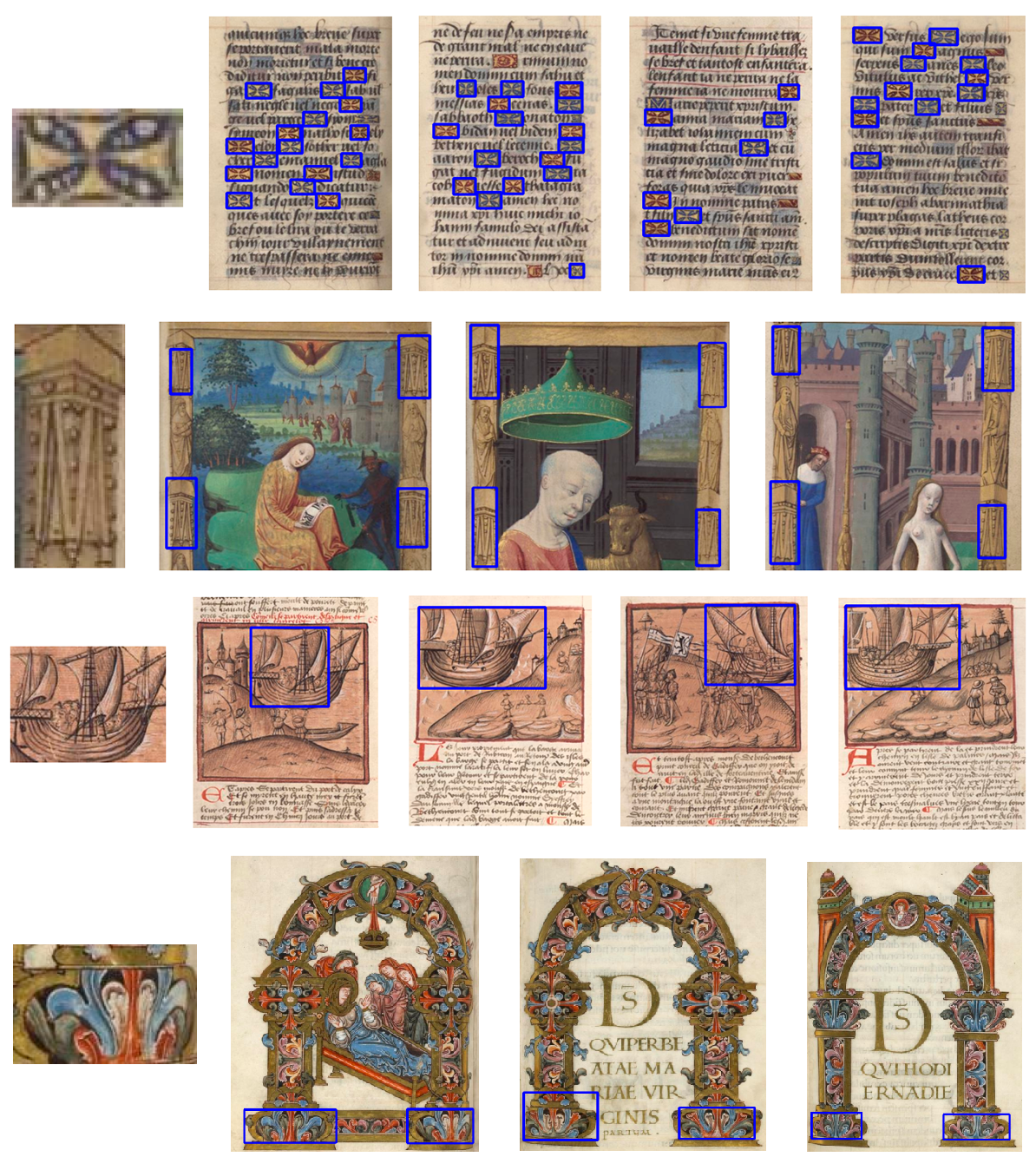}
     \caption{A sample of qualitative results of our proposal considering queries with diverse sizes and aspect ratios.}
     \label{fig:result_1}
\end{figure}

Figures  \ref{fig:result_1} illustrates some qualitative results of the proposed approach. We observe the effectiveness of our models for different sizes and aspect ratios of the queries.

%% file: conclusions.tex
\section{Conclusions}
This work presents a novel model for pattern spotting in historical documents, composed of two critical components. The first is a new visual encoder, iDoc, explicitly designed for medieval patterns. The second component is a semantic regions extractor implemented with the Grounding Dino model. We demonstrate that both elements are essential for achieving high performance and reducing processing time by 10x.

Our models outperform Curi's model for small, non-square patterns, which are the most challenging queries. For this queries, we have achieved a mAP of 0.588 in the pattern spotting task, representing an increase of $16.1\%$ and a gain of $37.7\%$ compared to the best-known model. Beyond pattern recognition, we demonstrated our model's competitive performance in document retrieval tasks. We achieved a mean Average Precision (mAP) of 0.789, 0.01 below the top models, but 10x faster.

Therefore, our new results could favor our proposed models for potential incorporation into real applications, facilitating the interaction of digitized historical documents.

\label{sec:conclusions}